\documentclass[letterpaper, 10 pt, conference]{ieeeconf}  

\IEEEoverridecommandlockouts                              

\usepackage{amsmath} 
\usepackage{amssymb}  

\usepackage{graphicx}
\usepackage{multirow}
\usepackage{threeparttable}
\usepackage{subfigure}

\usepackage{xcolor}
\definecolor{blizzardblue}{rgb}{0, 0.53, 0.91}
\newcommand{\ik}[1]{{#1}}
\newcommand{\px}[1]{{#1}}

\usepackage{blindtext}

\usepackage{comment}

\usepackage{microtype}
\usepackage[noadjust]{cite}

\title{\LARGE \bf
Human-Inspired Multi-Agent Navigation using Knowledge Distillation
}

\author{Pei Xu and Ioannis Karamouzas\\
\thanks{*This work was supported in part by the National Science Foundation under Grant No. IIS-2047632 and by an Amazon Research Award.}
\thanks{The authors are with the School of Computing at Clemson University, South Carolina, USA.
        {\tt\small \{peix,ioannis\}@clemson.edu}}%
}

\begin{document}

\maketitle
\thispagestyle{empty}
\pagestyle{empty}
\begin{abstract}

Despite significant advancements in the field of multi-agent navigation, agents still lack the sophistication and intelligence that humans exhibit in multi-agent settings. In this paper, we propose a framework for learning a human-like general collision avoidance policy for agent-agent interactions in fully decentralized, multi-agent environments. Our approach uses knowledge distillation with reinforcement learning  to shape the reward function based on expert policies extracted from human trajectory demonstrations through behavior cloning. We show that agents trained with our approach can take human-like trajectories in collision avoidance and goal-directed steering tasks not provided by the demonstrations, outperforming the experts as well as learning-based agents trained without knowledge distillation. 
\end{abstract}

\section{INTRODUCTION}
The problem of decentralized multi-agent navigation has been extensively studied in a variety of fields including robotics, graphics, and traffic engineering. 
Existing state-of-the-art planners 
can provide formal guarantees about the collision-free motion of the agents allowing applicability on physical robots~\cite{FS98,orca, alonso2013optimal}. 
In addition, recent learning-based approaches are capable of end-to-end steering and human-aware robot navigation~\cite{long2017deep,long2018towards,chen2019crowd}. 
Despite recent advancements, though, existing  agents cannot typically exhibit the level of sophistication and decision making that humans do in multi-agent settings. 
While the behavior of the agents can be improved through limited one-way communication~\cite{hildreth2019coordinating,godoy2020c}, and accounting for social norms~\cite{alahi2016social,chen2017socially,gupta2018social}, 
here we explore an alternative, data-driven approach that takes advantage of publicly available human trajectory datasets. 
Our approach starts with  expert demonstrations obtained from such trajectories and learns human-like navigation policies in fully decentralized multi-agent environments using reinforcement learning.

The idea of imitating expert behaviors is not something new. For example, behavior cloning techniques have been successfully applied to autonomous driving~\cite{bojarski2016end}, motion planning for autonomous ground robots~\cite{pfeiffer2017perception}, and 
distributed robot navigation~\cite{long2017deep} among others. However, pure imitation learning techniques 
are severely limited by the quality of the expert 
training dataset and cannot scale well to the multi-agent navigation domain due to its open-ended nature. 

Combining expert demonstrations with reinforcement learning can address the problem of insufficient training samples, with typical approaches relying on bootstrapping reinforcement learning with supervised learning~\cite{peters2008reinforcement,rajeswaran2017learning}, inverse reinforcement learning~\cite{ng2000algorithms}, and generative adversarial imitation learning~\cite{ho2016generative}. 
However such techniques are 
not directly applicable 
to the task of human-like collision avoidance learning 
since they typically assume reliable and representative expert demonstrations for the task in hand. 
Unfortunately, the experts (pedestrians) in human trajectory datasets are biased to some degree 
due to 
the fact that 
i) we only have access to a limited number of interaction data, which does not necessarily capture the behavior that the same expert will exhibit in a different setting;
and ii) trajectory datasets cannot capture the non-deterministic nature of human decision making, i.e., there are more than one trajectories that a human expert can take in the same setting.

To address these issues, we propose to use \emph{knowledge distillation}~\cite{hinton2015distilling} to learn a human-like navigation policy through expert policies extracted from human  
demonstrations. 
Given the imperfect nature of the experts, we avoid directly optimizing over them by adding an extra term to the objective function, as typically proposed in the literature~\cite{bertsekas2011approximate,nair2018overcoming}. 
Instead, we utilize the expert policies to shape the reward function during reinforcement learning
while promoting goal-directed and collision-free navigation. 
The resulting trained agents
can surpass the experts, and achieve better performance than 
pure learning-based agents without any expert reward signal and planning-based agents, while behaving in a human-like and more adept manner.

Overall, this paper makes the following contributions: 
\begin{enumerate}
\item We introduce a reinforcement learning approach for human-inspired multi-agent navigation that exploits human trajectory demonstrations and knowledge distillation to train a collision avoidance policy in homogeneous, fully decentralized settings. 
\item We experimentally show that the trained policy enables agents to take human-like actions for collision avoidance and goal-directed steering, 
and can generalize well to unseen scenarios not provided by the demonstrations, surpassing the performance of the experts  and that of pure reinforcement-learning based agents or planning-based agents.
\item We provide related code and pre-trained policies that can be used by the research community as baselines to facilitate further research and advance the field of human-inspired multi-agent navigation. 
\end{enumerate}

\section{RELATED WORK}

\subsection{Multi-Agent Navigation}
State-of-the-art techniques for decentralized multi-agent navigation can be broadly classified into local planning approaches and learning-based methods. 
Existing local planning approaches that rely on social forces, energy-based formulations, and rule-based techniques have been successfully applied to a variety of 
domains and  
have been shown to generate human-like collision avoidance behavior~\cite{RE99,helbing,prl}. 
In robotics, geometric local planners based on the concepts of velocity obstacles and time to collision~\cite{FS98,rvo,orca} are widely applicable as they 
provide formal guarantees about the collision-free behavior of
the agents 
and can be extended to account for motion and sensing uncertainty allowing implementation on
actual robots~\cite{alonso2013optimal,gorca,nhttc}. 
However, despite their robustness, local planning approaches typically require careful parameter tuning which can limit their applicability to unseen environments. 
In addition, such approaches typically rely on user-defined assumptions about the optimal motion principles that govern the interactions between the agents.

Learning-based decentralized methods typically employ a reinforcement learning paradigm to address the problem of limited training data, allowing agents to learn navigation policies through interactions with the environment~\cite{how,chen2017decentralized}. Such methods do not make assumptions explicitly about what the optimal policy should be, but rather let the agents learn that policy through trial and error based on a reward function. Despite lacking theoretical collision avoidance guarantees, such approaches allow for better generalization to new environments and conditions as compared to local planning methods, with some of the most recent works enabling   crowd-aware navigation for physical robots~\cite{how2,chen2019crowd,liu2020social} as well as fully distributed multi-robot navigation~\cite{long2018towards,fan2020distributed}. 
Our work is complementary to such learning-based methods, as we consider a reinforcement learning framework combined with imitation learning to train a human-like navigation policy applicable to homogeneous multi-agent navigation settings.

\subsection{Imitation Learning}
Below we consider imitation learning approaches that rely on limited number of expert demonstrations collected offline.
Under this assumption, prior work has focused on extracting an action policy to generate expert-alike controls.
Behavior cloning methods learn policies by directly matching the state-action pair as the input and output of the policy to the expert demonstrations through supervised learning~\cite{pomerleau1989alvinn,bojarski2016end}. 
Such approaches have also been explored for motion planning and distributed multi-robot navigation where expert demonstrations are based on simulations~\cite{pfeiffer2017perception,long2017deep}. 
Similar ideas are also applicable when the expert is represented by a distribution policy such that we can perform policy distillation to minimize the divergence or crossing entropy between the target policy and the expert policies~\cite{rusu2015policy,czarnecki2019distilling}. 
Inverse reinforcement learning (IRL)~\cite{ng2000algorithms}
methods estimate a parameterized reward function based on the expert trajectories and perform training using reinforcement learning. 
IRL has been successfully applied for robot navigation through crowded environments in~\cite{henry2010learning,kretzschmar2016socially,kim2016socially}. 
Generative adversarial imitation learning (GAIL)~\cite{ho2016generative}
have also been recently exploited for socially compliant robot navigation 
using raw depth images~\cite{tai2018socially}. 

While highly relevant, the aforementioned methods have difficulties when applied to the task of multi-agent human-like collision avoidance learning in general environments.
A typical assumption in imitation learning is that the expert policies or demonstrations are reliable. 
However, the demonstrations provided by pedestrian datasets are biased to some degree as a dataset can only contain a limited number of human-human interactions in certain situations. Also, there is a lot of uncertainty 
in human decision making 
which cannot be captured by trajectory datasets.  
To address these issues, we leverage the idea of knowledge distillation~\cite{hinton2015distilling},
and perform optimization implicitly through reward shaping during reinforcement learning based on the imperfect expert policies learned from human pedestrian trajectory demonstrations.

\section{APPROACH}
We propose a reinforcement learning framework for collision-avoidance and human-inspired multi-agent navigation. 
We assume a homogeneous, fully decentralized setup consisting of holonomic disk-shaped agents that do not explicitly communicate with each other but share the same action policy to perform navigation based on their own observations. 
To generate human-like actions, we take advantage of publicly available human trajectory data and perform reward shaping through knowledge distillation in the learning process.
Overall, our approach adopts a two-stage training process: 
(1) supervised learning on expert demonstrations from human trajectory data, and
(2) reinforcement learning with knowledge distillation to generate a general action policy.

\subsection{Problem Formulation}
We consider a decentralized multi-agent control setting, where each agent acts \emph{independently}, but all agents share the same action policy $\mathbf{a}_{i,t} \sim \pi_\theta(\cdot \vert \mathbf{o}_{i,t})$, 
with $\mathbf{a}_{i,t}$ denoting the action that the agent $i$ samples at a given time step $t$, $\mathbf{o}_{i,t}$ is the observation that the agent receives at $t$, and $\theta$ denotes the parameters of the policy.
The interaction process between an agent and the environment, which also includes other agents, is a partially observable Markov decision process, where each agent only relies on its own observations to make decisions.

For an $N$-agent environment, let $\mathcal{M}(\mathcal{S}_{t+1} \vert \mathcal{S}_{t}, \mathcal{A}_{t})$ be the transient model of the Markov decision process with state space $\mathcal{S}$ and action space $\mathcal{A}$. Then, 
at time $t$, 
$\mathbf{o}_{i,t}\in\mathcal{S}_{t}$ is a partial observable state for the $i$-th agent
and \mbox{$\mathcal{A}_t = \{\mathbf{a}_{i,t} \vert i=1,\cdots,N\}$}.
Given a time horizon $T$ and the trajectory $\tau = (\mathcal{S}_1, \mathcal{A}_1, \cdots, \mathcal{A}_{T-1}, \mathcal{S}_T)$, we optimize the parameter $\theta$ by maximizing the cumulative reward 
\begin{equation}\label{eq:reinforce}
J(\theta) = \mathbb{E}_{\tau \sim p_{\theta}(\tau)} \left[\mathcal{R}_t(\tau)\right]
\end{equation}
where $\mathcal{R}_t(\tau) =  \sum_{t=1}^{T-1}\sum_{i=1}^N\gamma^{t-1}r_{i,t}(\mathcal{S}_t, \mathbf{a}_{i,t}, \mathcal{S}_{t+1})$ is the total cumulative reward achieved by all agents with the step reward $r_{i,t}(\mathcal{S}_t, \mathbf{a}_{i,t}, \mathcal{S}_{t+1})$ and discount factor $\gamma$,
and $p_\theta(\tau)$ is the  state-action  visitation  distribution satisfying $p_\theta(\tau) = p(\mathcal{S}_{1})\prod_{t=1}^{T-1} \mathcal{M}(\mathcal{S}_{t+1}\vert\mathcal{S}_{t}, \mathcal{A}_{t})\prod_{i=1}^N\pi_\theta(\mathbf{a}_{i,t}\vert\mathbf{o}_{i,t})$.

Given that the decision process of each agent is completely independent under the fully decentralized context, the problem in Eq.~\ref{eq:reinforce} can be solved using a standard reinforcement learning setup 
by optimizing the combined cumulative reward of all agents based on their own trajectories, i.e:
\begin{equation}\label{eq:reinforce_decentralized}
J(\theta) = \frac{1}{N} \sum_{i=1}^N \mathbb{E}_{\tau_i \sim p_{\theta}(\tau_i)} \left[\sum_{t=1}^T \gamma^{t-1}r_{i,t}(\mathcal{S}_t, \mathbf{a}_{i,t}, \mathcal{S}_{t+1})\right]
\end{equation}
where $\tau_i = (\mathbf{o}_{i,1}, \mathbf{a}_{i,1}, \cdots, \mathbf{a}_{i,T-1}, \mathbf{o}_{i,T})$ is the observation-action trajectory for the $i$-th agent itself. 

\subsection{Observation and Action Space}\label{sect:spaces}
The observation space of each agent $i$ is defined using a local coordinate system based on the agent's current position, $\mathbf{p}_{i,t} \in \mathbb{R}^2$, and velocity, $\mathbf{v}_{i,t} \in \mathbb{R}^2$, as:
\mbox{$\mathbf{o}_{i,t} = \left[\mathbf{o}_{i,t}^{n_0}, \mathbf{o}_{i,t}^{n_1}, \cdots, \mathbf{o}_{i,t}^{n_m}, \mathbf{o}_{i,t}^g \mathbf{o}_{i,t}^v\right]$}, where
\begin{itemize}
\item $\mathbf{o}_{i,t}^{n_j} = [\mathbf{p}_{j,t} - \mathbf{p}_{i,t}, \mathbf{v}_{j,t} - \mathbf{v}_{i,t}]$ 
denotes the local state 
of the $j$-th neighbor. 
We assume that the agent has an observing radius $r$, and another agent $j$ is considered as its neighbor if $j \in \{j : \vert\vert \mathbf{p}_{j,t} - \mathbf{p}_{i,t} \vert\vert \leq r, \forall j \neq i\}$.
\item
 $\mathbf{o}_{i,t}^{n_0} = [\mathbf{0}, \mathbf{0}]$ is the neighborhood representation for the agent itself. This is a dummy neighbor representation to help data process when there is no neighbor in the observable range. 
    \item
$\mathbf{o}_{i,t}^g = \mathbf{g}_i - \mathbf{p}_{i,t}$ is the relative goal position where $\mathbf{g}_i \in \mathbb{R}^2$ is the goal position of the 
agent defined in the global coordinate system.
    \item
$\mathbf{o}_{i,t}^v = \mathbf{v}_{i,t}$ is the current agent's velocity.
\end{itemize}

The action space is a continuous 2D space denoting the expected velocity of the agent at the next time step, i.e.
$\mathbf{a}_{i,t} = \mathbf{\hat{v}}_{i,t+1}$. 
When applied on the agent, the expected velocity is scaled to satisfy the maximal speed $v_i^{max}$ at which agent $i$ can move: 
\begin{equation}
    \mathbf{v}_{i,t+1} = \begin{cases}
    \mathbf{\hat{v}}_{i,t+1} & \text{if } \vert\vert \mathbf{\hat{v}}_{i,t+1} \vert\vert \leq v_i^{max} \\
    v_i^{max}\mathbf{\hat{v}}_{i,t+1}/\vert\vert \mathbf{\hat{v}}_{i,t+1} \vert\vert & \text{otherwise.}
    \end{cases}
\end{equation}

\subsection{Reward}
We employ a reward function that gives 
a high reward signal $r_{arrival}$ as a bonus if an agent reaches its goal, and a negative reward signal $r_{collision}$ as a penalty to those that collide with any other agent.
These reward terms urge the agent to reach its 
goal in as few steps as possible while avoiding collisions. 
As opposed to introducing additional reward terms 
to regularize the agent trajectories, such as penalties for large angular speeds~\cite{long2018towards} and uncomfortable distances~\cite{liu2020social} or terms that promote social norms~\cite{chen2017socially},
our work mainly relies 
on human expert policies, $f_{e}(\mathbf{o}_{i, t})$, extracted from real crowds trajectories to shape the reward. 

In particular, we seek for the agents to 
exhibit navigation behavior in the style of the
experts without 
explicitly dictating their behaviors. 
However, expert policies obtained 
by learning from human trajectory demonstrations are typically not general enough to lead agents to effectively resolving collisions in unseen environments. 
As such, we avoid directly optimizing the action policies through cloning behaviors from imperfect 
experts by introducing auxiliary loss terms in the objective function (Eq.~\ref{eq:reinforce_decentralized}). 
Instead, we propose to use \emph{knowledge distillation}   
and exploit the behavior error $\frac{1}{K}\sum_{e=1}^{K}\vert\vert \mathbf{a}_{i,t} - f_{e}(\mathbf{o}_{i, t}) \vert\vert$ averaged over $K$ expert policies to shape the \emph{reward function}. 

During inference, expert policies would become unreliable
if the agent's speed is significantly different from the demonstrations. 
Humans typically prefer to walk at a certain \emph{preferred speed} depending on the environment and their own personality traits,
and their behavior patterns may differ a lot at different walking speed.
As such, 
we introduce a velocity regularization term in the reward function to encourage each agent to move at a preferred speed
$v^\ast$
similar to the ones observed in the demonstrations. 
To promote goal seeking behavior and avoid meaningless exploration led by imperfect experts, $v^\ast$ is used to scale the goal velocity that points from an agent's current position to its goal. 

Given an environment with $N$ agents, the complete reward function 
of our proposed imitation learning approach with knowledge distillation from $K$ experts is defined as follows
\begin{multline}\label{eq:reward}
     r_{i,t}(\mathcal{S}_t, \mathbf{a}_{i,t}, \mathcal{S}_{t+1}) = \\
    \begin{cases}
    r_{arrival} & \text{if } \vert\vert \mathbf{g}_i - \mathbf{p}_{i,t} \vert\vert \leq R \\
    r_{collision} & \text {if } \vert\vert \mathbf{p}_{j, t+1} - \mathbf{p}_{i, t+1} \vert\vert \leq 2R \\
    w_e r_{i,t}^e + w_v r_{i,t}^v & \text{otherwise}
    \end{cases}
\end{multline}
where $j = 1, \cdots, N$ with $j \neq i$, $R$ is the agent radius, and $w_v$ and $w_e$ are weight coefficients. The knowledge distillation reward term $r_{i,t}^e$ and the velocity regularization term $r_{i,t}^v$ are 
computed as: 
\begin{equation}\label{eq:reward_terms}\begin{array}{c}
     r_{i,t}^e = \exp{\left(-\frac{\sigma_{e}}{K}\sum_{e=1}^{K} \vert\vert \mathbf{a}_{i,t} -  f_{e}(\mathbf{o}_{i, t}) \vert\vert\right)},\\
     r_{i,t}^v = \exp{\left(-\sigma_{v} \vert\vert \mathbf{v}_{i,t+1} - \mathbf{v}_{i,t+1}^\ast \vert\vert\right)}
\end{array}
\end{equation}
where $\sigma_{e}$ and $\sigma_{v}$ are scale coefficients, and $\mathbf{v}_{i,t+1}^\ast = v^\ast(\mathbf{g}_i - \mathbf{p}_{i, t})/\vert\vert \mathbf{g}_i - \mathbf{p}_{i, t} \vert\vert$ is the goal velocity of the agent $i$ having a magnitude equal to the preferred speed $v^\ast$. 

\subsection{Policy Learning}\label{sect:learning}
The learning process of our approach has two stages: (1) obtain expert policies by supervised learning from real human pedestrian trajectories, and (2) train a shared action policy in multi-agent environments through reinforcement learning to maximize Eq.~\ref{eq:reinforce_decentralized} with the reward function defined in Eq.~\ref{eq:reward}.

\vspace{0.1cm}
\noindent{\emph{Supervised Learning. }}
Expert policies are trained by supervised learning on human pedestrian trajectory datasets
where the observation-action training data $ \{(\mathbf{o}_{i,t}, \mathbf{a}_{i,t})\}$ is extracted as  described in Section~\ref{sect:spaces}.
The goal position is defined by the last position of each pedestrian appearing in the datasets.
We use a neural network denoted by $f_e(\cdot\vert\phi_e)$ with parameters $\phi_e$ to perform action prediction, which is optimized by minimizing the mean squared error between the predicted and the ground truth action:
\begin{equation}\label{eq:supervise_loss}
    \mathcal{L}_{sup} = \mathbb{E}_{(\mathbf{o}_{i,t}, \mathbf{a}_{i,t})} \left[\| f_e(\mathbf{o}_i\vert\phi_e) - \mathbf{a}_i \|^2 \right] 
\end{equation}

Trajectory data in human datasets is typically 
recorded by sensors at fixed and rather sparse time intervals.  
Training on such limited, discrete data is easily prone to overfitting,
as the network simply remembers all the data. 
To alleviate this issue, we perform data augmentation during training by 
randomly sampling a continuous timestamp and using linear interpolation to estimate corresponding observation-action data from discrete trajectories.
In addition to linear interpolation, 
we adopt two more types of data augmentation: scenario flipping and rotation. 
During training, the x- and/or y-axis of each observation-action datum 
is flipped by a stochastic process 
along with randomly rotating the observation-action local system. 
This helps increase the generality of the training samples
without destroying the relational information between human-human interactions captured in the dataset.  

While multiple expert policies can be trained by using different datasets, 
the action patterns of pedestrians 
recorded at different times and/or under different scenarios may vary a lot. 
This could lead to too much divergence when we ensemble the expert policies 
in the reward function of the reinforcement learning stage (cf. Eq~\ref{eq:reward}). 
As such, in our implementation, 
we extract expert policies only from one pedestrian trajectory dataset.
Under this setting, 
the deviation caused by stochastic factors during expert training can be effectively eliminated by data augmentation (Section~\ref{sect:sensitive}). 
This allows us to employ just a single expert policy in all of our experiments and obtain a  final action policy that is comparable to the one obtained with multiple experts but at a significant training speedup.



\vspace{0.1cm}
\noindent{\emph{Reinforcement Learning:}} 
To perform reinforcement learning, we exploit the DPPO algorithm~\cite{heess2017emergence}, which is a distributed training scheme that relies on the PPO algorithm~\cite{schulman2017proximal}. 
DPPO 
optimizes Eq.~\ref{eq:reinforce_decentralized} using policy gradient method~\cite{sutton2000policy} by maximizing
    $J(\theta) = \frac{1}{N}\sum_i^N \mathbb{E}_t\left[\log \pi_\theta(\mathbf{a}_{i,t} \vert \mathbf{o}_{i,t}) \hat{A}_{i,t}\right]$,
where $\hat{A}_{i,t}$ is an estimation to 
the discounted cumulative reward term
with bias subtraction, which in our implementation is computed by the generalized advantage estimation (GAE)~\cite{schulman2015high}. 

To help exploration, we also introduce differential entropy loss to encourage exploration and avoid premature convergence, resulting the following objective function: 
\begin{equation}
\label{eq:reinforce_loss}
    \mathcal{L}_{re}(\theta) = \frac{1}{N}\sum_i^N\mathbb{E}_t\left[\log\pi_\theta(\mathbf{\bar{a}}_{i,t} \vert \mathbf{\bar{o}}_{i,t}) \hat{A}_{i,t}
    + \beta \mathcal{H}(\pi_{\theta}(\cdot\vert\mathbf{\bar{o}}_{i,t}))\right]
\end{equation}
where $\mathcal{H}$ represents the entropy of the action policy, 
and 
$\beta$ is a scalar 
coefficient of the policy entropy loss. 
The parameters 
$\mathbf{\bar{o}}_{i,t}$ and $\mathbf{\bar{a}}_{i,t}$ denote the observation and action, respectively, after performing goal direction alignment, i.e. 
after 
rotating the local coordinate system described in~\ref{sect:spaces} such that the goal 
is always at a fixed direction (the positive direction of x axis in our implementation). 
By this method, 
we can remove one dimension of $\mathbf{o}_{i,t}^g$ and
use the distance to the goal position instead, i.e., $\bar{o}_{i,t}^g = \vert\vert \mathbf{g}_i - \mathbf{p}_{i,t}\vert\vert$. 
This trick helps the learning,
as it reduces the complexity of state space and is beneficial to exploration. 
Intuitively, agents in the goal-aligned local system are more likely to encounter similar observations and thus exploit previous experience more effectively.

We do not apply the goal-alignment trick during expert policy training, 
since it could result in too much overfitting.
In human trajectory datasets, pedestrians move mainly towards their goals.
With the alignment trick the output actions (velocities) would mostly respond to the goal direction.     
As a result, expert policies 
would prefer orientation adaption more than speed adaptation 
which can often lead to understeering. 

\begin{figure}[tb]
\centering
\includegraphics[width=0.6\linewidth]{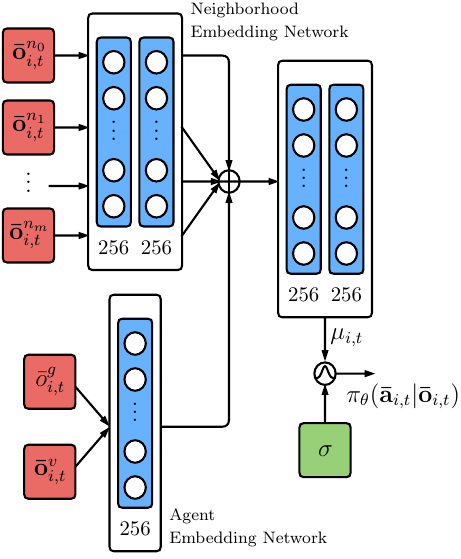}
\caption{Network architecture employed as the policy network during reinforcement learning. The policy is defined as a Gaussian distribution with $\mu_{i,t}$ as the mean value and an input-independent parameter $\mathbf{\sigma}$ as the standard deviation. The same architecture is adopted for the value network 
as well as the expert policy learning.}
\label{fig:network}
\end{figure}
  
\noindent\emph{Network architecture. }
We use a bivariate Gaussian distribution with independent components as the action policy,
where the mean values are provided by the policy network with architecture shown in Fig.~\ref{fig:network},
and the standard deviation is a 2-dimension observation-independent parameter.
Neighbor representations, 
$\mathbf{\bar{o}}_{i,t}^{n_0}$ and $\mathbf{\bar{o}}_{i,t}^{n_1}, \cdots, \mathbf{\bar{o}}_{i,t}^{n_m}$, 
are simply added together with the agent representation, 
$[\bar{o}_{i,t}^g, \mathbf{\bar{o}}_{i,t}^v]$, 
after the embedding networks. 
$\mathbf{\bar{o}}_{i,t}^{n_0} = \mathbf{o}_{i,t}^{n_0} = [\mathbf{0}, \mathbf{0}]$
denotes the neighborhood representation of the agent itself and is always kept 
such that the network can work normally when there is no neighbors observed. 
This network architecture can support observations 
with an arbitrary number of neighbors 
and the 
embedding result 
does not rely on the input order of the neighbor representation sequence. 
We use the same architecture for the value network to perform value estimation for the GAE computation, 
and to train the expert policies through supervised learning without goal alignment observations.  

\begin{table*}[!tb]
\caption{Quantitative Evaluation}
\label{table:quat}
\begin{center}
\begin{threeparttable}[t]
\begin{tabular}{l|l|c|c|c|c|c|c}
\hline
& 
& 20-Circle
& 24-Circle
& 20-Corridor 
& 24-Corridor
& 20-Square
& 24-Square \\
\hline
& ORCA & $0.91\pm0.15$ & $0.71\pm0.17$ & $\mathbf{1.00\pm0.00}$ & $\mathbf{1.00\pm0.00}$ & $\mathbf{1.00\pm0.00}$ & $\mathbf{1.00\pm0.00}$\\
Success Rate
& SL & $0.40\pm0.17$ & $0.23\pm0.12$ & $0.64\pm0.15$ & $0.59\pm0.14$ & $0.64\pm0.15$ & $0.58\pm0.13$\\
{\scriptsize (higher is better)}
& RL & $0.97\pm0.05$ & $0.96\pm0.05$ & $0.93\pm0.08$ & $0.92\pm0.08$ & $0.91\pm0.09$ & $0.88\pm0.11$\\
& Ours & $\mathbf{1.00\pm0.01}$ & $\mathbf{0.99\pm0.02}$ & $\mathbf{1.00\pm0.00}$ & $0.99\pm0.03$ & $0.99\pm0.03$ & $0.97\pm0.06$\\


\hline
& ORCA & $\mathbf{0.17\pm0.33}$ & $\mathbf{0.27\pm0.41}$ & $0.13\pm0.47$ & $0.20\pm0.71$ & $0.23\pm0.49$ & $0.40\pm0.80$\\
Extra Distance
& SL & $3.80\pm4.67$ & $6.37\pm6.39$ & $6.06\pm8.49$ & $7.29\pm9.32$ & $5.83\pm7.85$ & $6.92\pm9.15$\\
{\scriptsize (lower is better)}
& RL & $0.81\pm0.69$ & $0.93\pm0.73$ & $0.25\pm0.44$ & $0.31\pm0.53$ & $0.43\pm0.61$ & $0.48\pm0.59$\\
& Ours & $0.34\pm0.30$ & $0.49\pm0.29$ & $\mathbf{0.09\pm0.08}$ & $\mathbf{0.01\pm0.09}$ & $\mathbf{0.03\pm0.13}$ & $\mathbf{0.04\pm0.12}$\\

\hline
& ORCA & $0.34\pm0.11$ & $0.27\pm0.17$ & $0.35\pm0.07$ & $0.35\pm0.09$ & $0.32\pm0.10$ & $0.30\pm0.11$\\
Energy Efficiency
& SL & $0.06\pm0.20$ & $0.04\pm0.20$ & $-0.02\pm0.20$ & $-0.02\pm0.19$ & $-0.02\pm0.20$ & $-0.04\pm0.19$\\
{\scriptsize (higher is better)}
& RL & $0.29\pm0.05$ & $0.28\pm0.06$ & $0.30\pm0.04$ & $0.30\pm0.05$ & $0.29\pm0.06$ & $0.29\pm0.06$\\
& Ours & $\mathbf{0.36\pm0.05}$ & $\mathbf{0.36\pm0.05}$ & $\mathbf{0.36\pm0.07}$ & $\mathbf{0.37\pm0.05}$ & $\mathbf{0.36\pm0.06}$ & $\mathbf{0.36\pm0.05}$\\

\hline
\end{tabular}
\end{threeparttable}
\end{center}
\footnotetext[1]{Footnote}
\end{table*}

\section{EXPERIMENTS}
\subsection{Simulation Environment Setup}

Our simulation environment has three scenarios shown in Fig.~\ref{fig:scenarios}. 
In the circle scenarios, agents are roughly placed on the circumference of a circle and have to reach their antipodal positions; 
in the corridor scenarios, agents are randomly initialized at 
the two sides,
oriented either vertically or horizontally, and have random goals at the opposite sides; 
and in the square crossing scenarios, agents are placed at the four sides 
and need to walk across the opposite sides to reach randomly placed goals. 
To increase the generality, all scenarios are generated randomly during simulation. 
The circle scenario has a random radius varying from 4m to 6m.
The other two scenarios have widths and heights in the range of 8m to 12m. 

During each simulation, 6 to 20 agents, sharing the same action policy under optimization, are 
placed into the scenarios and are given stochastic goal positions. 
The simulation time step size is 0.04s, while agents receive control signals every 0.12s. 
All agents are modeled as disks with a radius of 0.1m,
and have a preferred speed $v^\ast$ of 1.3m/s with a maximum speed limit of 2.5m/s to imitate human walking patterns 
in the dataset for expert policy training.
During simulation, collided agents are kept as static obstacles to other agents, while agents that arrive at their goals are removed from the environment. 
Each simulation episode terminates if all non-collided agents reach their goals or a time limit of 120s is met. 


\subsection{Training Details}
We exploit the \textit{students} dataset~\cite{lerner2007crowds} for supervised learning of expert policies. 
This dataset records the trajectories of 434 pedestrians in an open campus environment, 
providing a more general setting to learn typical expert policies as compared to environments in other publicly available datasets consisting of too many obstacles and/or terrain restrictions. 
Each pedestrian is considered as a disk agent with a radius of 0.1m, estimated by the typical minimal distance between trajectories; 
the goal position is chosen as the last 
position of each pedestrian's trajectory; 
velocities are estimated by finite differences. 

Before training, we preprocess the raw data to remove 
pedestrians that stand still or saunter without clear goals, as trajectories of such pedestrians cannot help learn navigation policies and would become training noise.
After data cleansing, we kept the observation-action records from 300 out of the 434 pedestrians for expert training (Eq.~\ref{eq:supervise_loss}),
while the rest of the pedestrian trajectories are only used as part of the neighborhood representation of the active pedestrians. 
During the following experiments, we, by default, use only one expert policy trained with data augmentation (see Section~\ref{sect:sensitive} for related analysis). 
The default values of the reward function weights are: $w_e=0.08$, $w_v = 0.02$, $\sigma_e = \sigma_v = 0.85$.
We refer to \verb|https://github.com/xupei0610/KDMA| for all hyperparameters used, along with related videos and code. 

\subsection{Quantitative Evaluation}
\begin{figure}[tb]
\centering
\includegraphics[width=.9\linewidth]{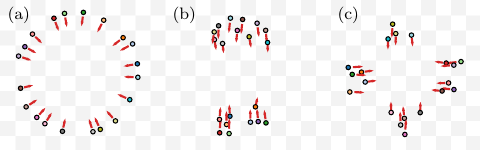}
\caption{Our environment consists of: (a) a circle crossing scenario, 
(b) a corridor scenario, 
and (c) a square crossing scenario.  
During each simulation episode, a scenario is randomly chosen and 6 to 20 agents are put into the scenario randomly. 
Red arrows indicate goal directions.}
\label{fig:scenarios}
\end{figure}

We compare our method to three approaches: the geometric-based 
method of Optimal Reciprocal Collision Avoidance (ORCA)~\cite{orca}, 
a supervised learning approach (SL), and a reinforcement learning approach without knowledge distillation (RL). 
To prevent ORCA agents from staying too close to each other, we increase the agent radius by 20\% in the ORCA simulator. 
SL denotes the performance of the expert policy obtained in our approach. 
RL uses the reward function from~\cite{long2018towards}, which optimizes agents to reach their goals as fast as possible without collisions. 
Since RL performs optimization based on the maximal agent speed but  our performance evaluation are computed based on a preferred speed of 1.3m/s, for fairness, 
we set the maximal speed in RL training and testing cases as 1.3m/s instead of the default value of 2.5m/s.

We perform 50 trials of testing, which are the same across different methods, using the three scenarios shown in Fig.~\ref{fig:scenarios}. 
We run comparisons using three evaluation metrics 
and report the results in Table~\ref{table:quat}. Reported numbers are the mean $\pm$ std. 
\textit{Success rate} denotes the percentage of agents that reached their goals.
\textit{Extra distance} is the additional distance in meters that an agent traversed instead of taking a straight line path  to the goal. 
This metric is recorded only for agents that reached their goals.
\textit{Energy efficiency} measures the ratio between an agent's progress towards its goal and its energy consumption at each time step~\cite{godoy2020c}. The energy consumption is computed following the definition in~\cite{godoy2020c} such that the optimal speed in terms of energy expended per meter traversed is 1.3m/s. 

As shown in Table~\ref{table:quat}, our approach can achieve as good as or better results than ORCA in terms of success rate
and energy efficiency, while it  significantly outperforms RL and SL in most of the scenarios. Given that the expert policies obtained by SL are used in the second stage of our approach, it is clear that  
agents with knowledge distillation can surpass the experts to achieve collision-free and energy-efficient behavior patterns. 
Regarding the extra distance traversed, agents with our approach favor shortest paths in the corridor and square scenarios. In the circle scenarios, ORCA 
outperforms the  methods as its agents prefer to 
take straight line paths to the goal that passes through the center of the environment by mainly adapting their speeds.  
However, as discussed below these are not typically the types of trajectories that humans would exhibit.
In addition, we note that the 
behavior of ORCA agents varies a lot depending on the choices of parameters used, including the size of the simulation time step and the safety buffer added to the radius of the agent. 
\subsection{Comparisons to Human Reference Trajectories}


\begin{figure}[tb]
\centering
\includegraphics[width=.23\linewidth]{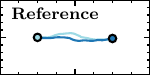}
\includegraphics[width=.23\linewidth]{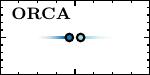}
\includegraphics[width=.23\linewidth]{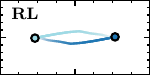}
\includegraphics[width=.23\linewidth]{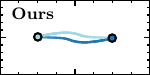}
\caption{Trajectories of different methods in 2-agent head-to-head scenario.
}
\label{fig:traj_s2}
\end{figure}

\begin{figure}[tb]
\centering
\includegraphics[width=.9\linewidth]{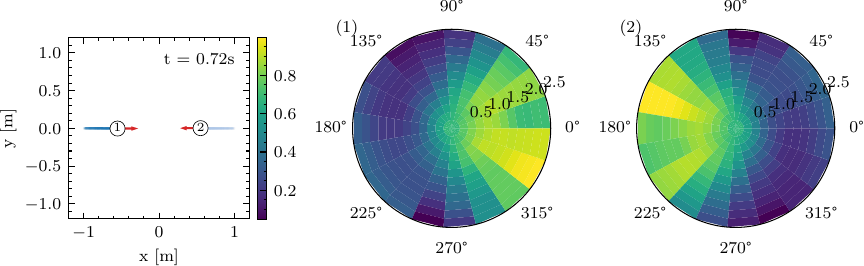}
\includegraphics[width=.9\linewidth]{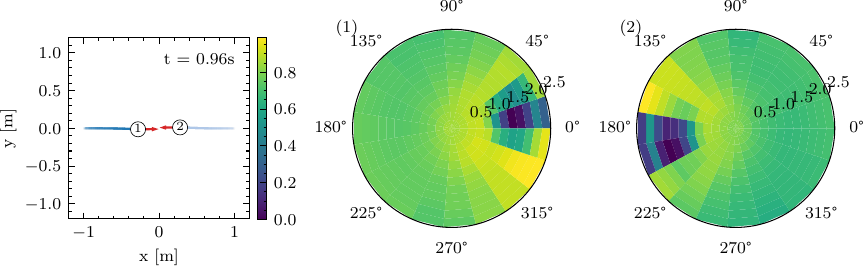}
\caption{Value heatmaps of our approach in the 2-agent interaction scenario. Brighter radial bins indicate the action decisions that are considered better by the agents. Dark bins indicate the actions that agents do not prefer.}
\label{fig:heatmap_s2}
\end{figure}

\begin{figure}[tb]
\centering
\includegraphics[width=.9\linewidth]{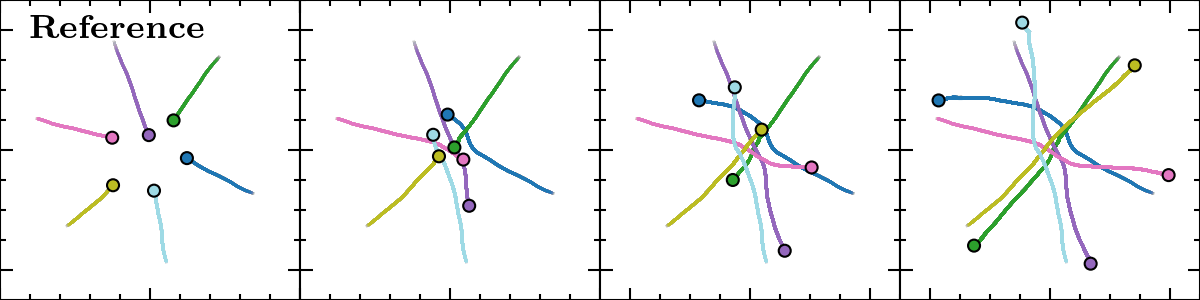}
\includegraphics[width=.9\linewidth]{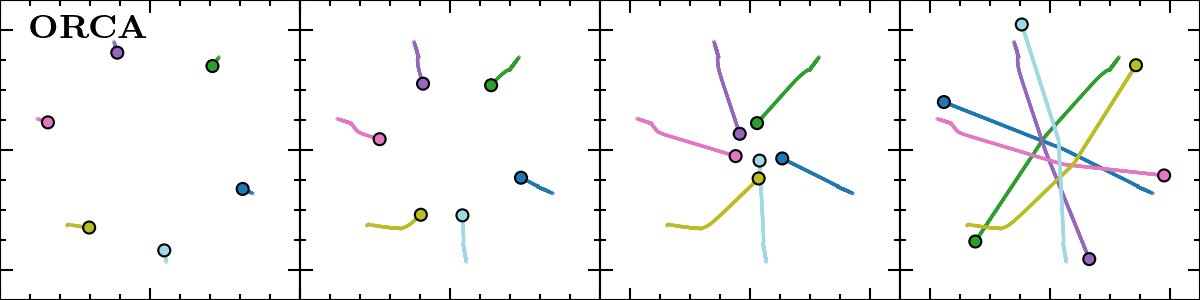}
\includegraphics[width=.9\linewidth]{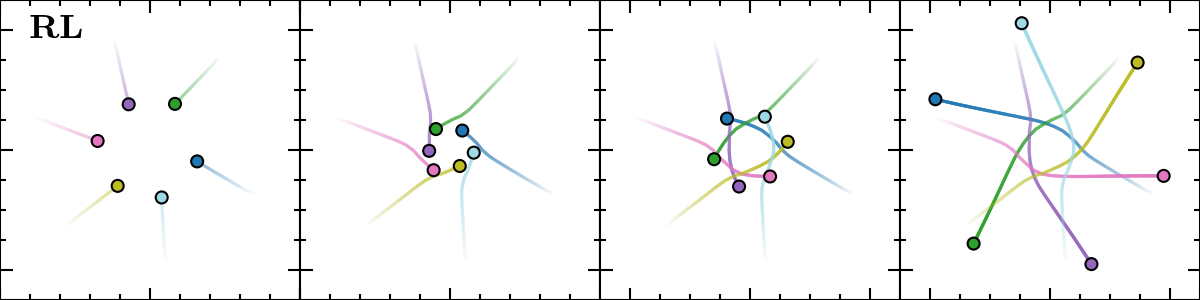}
\includegraphics[width=.9\linewidth]{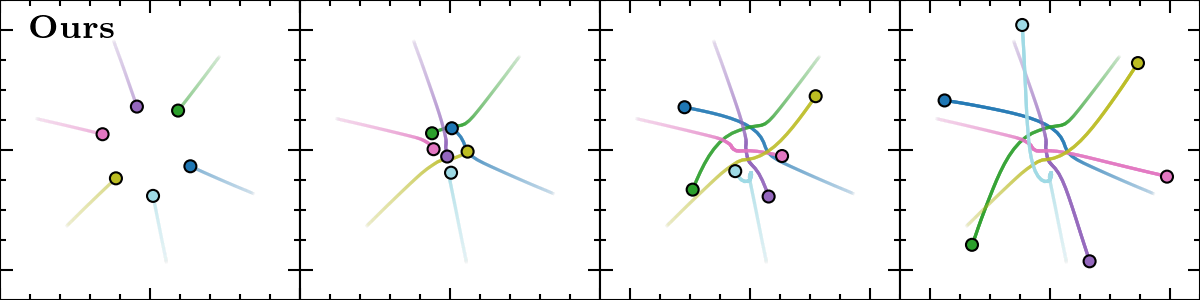}
\caption{Agent trajectories in a 6-agent circle scenario. Each column is captured at the same time fraction of the whole trajectories.}
\label{fig:traj_c6_ref}
\end{figure}

Figure~\ref{fig:traj_s2} compares the trajectories generated by different methods in a symmetric head-to-head interaction scenario to 
 reference data obtained from motion captured human experiments~\cite{moussaid}. 
Due to the symmetric nature of the scenario, ORCA agents fail to reach their goals and end up standing still next to each other in the middle of the environment. 
RL agents maintain a high walking speed close to their maximal one and solve the interaction by performing an early orientation adaptation as opposed to the reference data where the avoidance happens later on. 
Trajectories generated by our approach more closely match the reference trajectories. 
To further analyze the decision making of the agents in our approach,
Fig.~\ref{fig:heatmap_s2} depicts the corresponding action value heatmaps 
obtained by estimation through the value network of DPPO.
As shown, at 0.72s the agents implicitly negotiate to pass each other from the right side; and later on at the 0.92s mark every action that can lead to an imminent collision is forbidden to both agents, with agent 1 preferring to turn a bit to the right while moving at a speed of at least 1m/s while agent 2 decides to veer a bit to the right and accelerate.

Figure~\ref{fig:traj_c6_ref} shows trajectories from a 6-agent circle scenario, where the reference human trajectories were obtained from ~\cite{wolinski2014parameter}. 
In this scenario, ORCA agents mainly resolve collisions by adapting their speeds; they move slow at the beginning due to the uncertainty about what their neighbors are doing and eventually do a fast traverse mostly along straight lines towards their goals. 
RL agents, on the other hand, prefer to travel at their maximum speed towards the center of the environment, and then adapt their orientations in unison resulting in a vortex-like avoidance pattern.
Agents in our approach exhibit more variety 
by using a combination of both speed and orientation adaptation to balance the tradeoff between collision avoidance and goal steering, 
As such, the resulting trajectories can capture to some degree the diversity 
that humans exhibit in the reference data. 

\begin{figure}[tb]
\centering
\includegraphics[width=\linewidth]{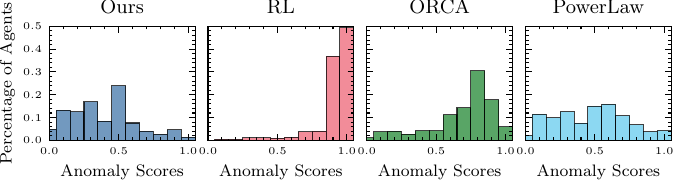}
\caption{Distributions of anomaly scores on \textit{students} scenario.
}
\label{fig:novelty}
\end{figure}

\begin{figure}[tb]
\centering
\includegraphics[width=\linewidth]{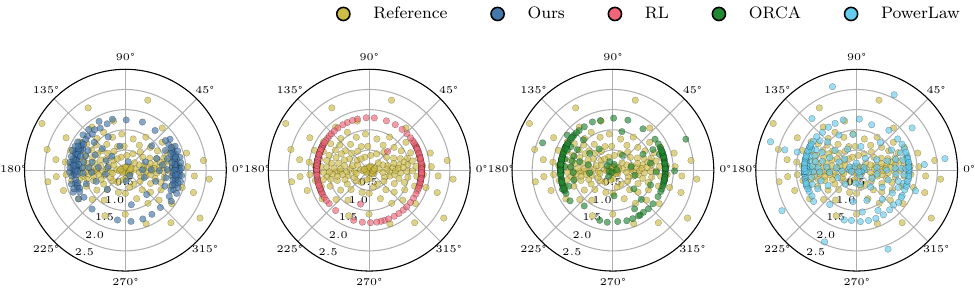}
\caption{Agent velocity distributions on \textit{students} scenario.}
\label{fig:vel_dist}
\end{figure}

To statistically analyze the quality of generated trajectories, 
we reproduce the \textit{students} scenario with different methods 
and run novelty detection using the $k$-LPE algorithm~\cite{zhao2009anomaly}. 
\ik{As a comparison, we also introduce another baseline, the PowerLaw method~\cite{prl}, which is a state-of-the-art crowd simulation method elaborated from the analysis of pedestrian datasets including the \textit{students} one.} 
Figure~\ref{fig:novelty} shows how the corresponding agent trajectories compare to the trajectories in the \textit{students} reference dataset based on the metrics of the agent's linear speed, angular speed, and distance to the nearest neighbor.
As it can be seen, 
the majority of trajectories generated by our approach have low anomaly scores as compared to ORCA and RL. 
Figure~\ref{fig:vel_dist} 
further highlights that agents 
in our approach 
can more closely match the velocity distribution of the ground truth data. 
\px{We note that PowerLaw provides high fidelity simulation as well. However, to guarantee numerical stability, a very small simulation time step is typically required during each sensing-acting cycle,  which may not always be applicable to real robots.  
}

\subsection{Sensitivity Analysis}\label{sect:sensitive}

\begin{figure}[!t]
\centering
\subfigure[Performance with different expert policies.]{
\label{fig:sen_aug}
\includegraphics[width=.9\linewidth]{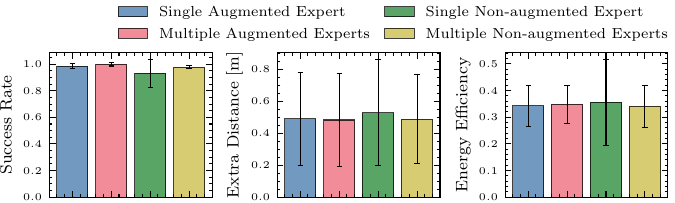}
}
\subfigure[Performance with different reward terms.]{
\label{fig:sen_rew}
\includegraphics[width=.9\linewidth]{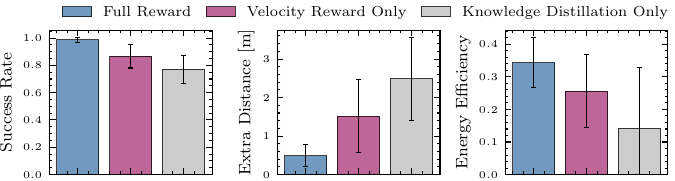}
}
\caption{Sensitivity analysis based on performance in the 24-agent circle. (a) Multiple experts testing cases use three experts simultaneously for reward computation. Single expert results  are the average of three training trials each of which uses only one of the corresponding three experts. (b) Comparisons of different reward terms using a single expert with data augmentation. }
\label{fig:sensitivity}
\end{figure}

Figure~\ref{fig:sen_aug} shows the performance of action policies trained with different experts 
in the 24-Circle scenario. 
As it can be seen, the performance varies a lot  when employing a single expert trained without data augmentation. However, the uncertainty caused by stochastic factors of supervised training can be effectively eliminated  by introducing data augmentation during expert policy training. 
Using either multiple augmented experts or a single expert with data augmentation results in  comparable 
performances. 
As such, and taking into account the fact that the inference time increases dramatically as more expert policies are exploited simultaneously, we employed  a single expert policy with data augmentation in all of our experiments. 
In our tests, using two V100 GPUs, it took approximately six hours to perform training with three expert policies,
while training with a single expert took only around three and a half hours.

Figure~\ref{fig:sen_rew} compares the training results obtained by using the full reward  function that considers both the knowledge distillation term and the velocity regularization term  (Eq.~\ref{eq:reward_terms})  
to results obtained by using only one of the two terms.  
As shown in the figure, the expert policy by itself is not quite reliable, 
as it leads to a worse performance across all three metrics when only the knowledge distillation term is used. 
However, when combined with the velocity reward term,  
the expert can help agents achieve the best overall performance. 

\section{LIMITATIONS and FUTURE WORK}

We propose a framework for learning a human-like general collision avoidance policy for agent-agent interactions 
in fully decentralized multi-agent environments.
To do so, we use knowledge distillation implicitly in a reinforcement learning setup to shape the reward function based on expert policies extracted from human pedestrian trajectory demonstrations.
Our approach can help agents surpass the experts, and
achieve better performance and more human-like action patterns, compared to using reinforcement learning without knowledge distillation and to existing geometric planners. 
\px{
During testing, agents use the trained navigation policy deterministically.
In the future, to account for the uncertainty of human navigation decisions, we would like to test stochastic actions by keeping the executed policy as a Gaussian distribution instead of using only the mean value.
}


A limitation of our approach is that 
we assume all agents are of the same type, and particularly are holonomic disk-shaped agents.
While, in theory, the same trained policy can be ported to other types of agents, 
the resulting behavior can be very conservative as the true geometry of the target agent will be ignored during training. 
In addition, we assume 
that trained agents can behave similarly to the expert  pedestrians that provide the demonstrations,  
ignoring the kinodynamic constraints of specific robot platforms. 
Even though this issue can be addressed by relying on a controller to convert human-like velocity commands to low-level robot control inputs, 
it also opens an interesting direction for future work that focuses on mining robot-friendly action patterns rather than action commands from human  trajectories. 
Another avenue for future work is extending our method to mixed settings where agents can interact with humans. 
The recent works  of~\cite{gupta2018social,liu2020social,sathyamoorthy2020densecavoid} on socially-aware robot navigation 
and of~\cite{mavrogiannis2018social,mavrogiannis2018multi} on generating socially compliant behaviors in multi-agent environments 
can provide some interesting ideas towards this research direction.

\bibliography{IEEEabrv.bib,root.bib}{}
\bibliographystyle{IEEEtran}

\end{document}